\documentclass{article}

\usepackage[utf8]{inputenc} %
\usepackage[T1]{fontenc}    %
\usepackage{hyperref}       %
\usepackage{url}            %
\usepackage{booktabs}       %
\usepackage{amsfonts}       %
\usepackage{nicefrac}       %
\usepackage{microtype}      %

\usepackage{bm}
\usepackage{dsfont}
\usepackage{amsmath,amssymb}
\usepackage{cite}
\usepackage{color}
\usepackage{soul}
\usepackage{algorithm,algorithmic}
\usepackage{wasysym}
\usepackage{float}
\usepackage{xfrac}

\newcommand{\gen}{\mathcal{G}(\vz; \vtheta)}

\newcommand{\tdist}{p^\ast\!} %
\newcommand{\tdistx}{\tdist(\vx)}
\newcommand{\zdist}{q(\vz)} 
\newcommand{\qtheta}{q_{\theta}}
\newcommand{\qthetax}{q_{\theta}(\vx)} 
\newcommand{\rphi}{r_{\phi}} 
\newcommand{\rphix}{r_{\phi}(\vx)} 
\newcommand{\rstarx}{r^*(\vx)}

\newcommand{\LBregman}{\mathcal{L}_{B}}

\newcommand\cut[1]{}

\newcommand{\squishlist}{
   \begin{list}{$\bullet$}
    { \setlength{\itemsep}{0pt}      \setlength{\parsep}{3pt}
      \setlength{\topsep}{3pt}       \setlength{\partopsep}{0pt}
      \setlength{\leftmargin}{1.5em} \setlength{\labelwidth}{1em}
      \setlength{\labelsep}{0.5em} } }

\newcommand{\squishlisttwo}{
   \begin{list}{$\bullet$}
    { \setlength{\itemsep}{0pt}    \setlength{\parsep}{0pt}
      \setlength{\topsep}{0pt}     \setlength{\partopsep}{0pt}
      \setlength{\leftmargin}{2em} \setlength{\labelwidth}{1.5em}
      \setlength{\labelsep}{0.5em} } }

\newcommand{\squishend}{
    \end{list}  }

\newcommand{\JSpq}[2]{\mbox{JS}\left[{#1}||{#2}\right]}
\newcommand{\JSpqpi}[2]{\mbox{JS}_{\pi}\left[{#1}||{#2}\right]}

\newcommand{\KL}{\mbox{KL}}

\newcommand{\myvec}[1]{\mathbf{#1}}
\newcommand{\myvecsym}[1]{\boldsymbol{#1}}

\newcommand{\vphi}{\myvecsym{\phi}}

\newcommand{\vtheta}{\myvecsym{\theta}}

\newcommand{\vx}{\myvec{x}}

\newcommand{\vy}{\myvec{y}}

\newcommand{\vz}{\myvec{z}}

\newcommand{\E}{\mathbb{E}}

\newcommand{\be}{\begin{equation}}
\newcommand{\ee}{\end{equation}}
\newcommand{\bea}{\begin{eqnarray}}
\newcommand{\eea}{\end{eqnarray}}
\newcommand{\beaa}{\begin{eqnarray*}}
\newcommand{\eeaa}{\end{eqnarray*}}

\DeclareMathAlphabet{\mathpzc}{OT1}{pzc}{m}{n}

\usepackage{balance}

\usepackage{times} %
\usepackage[accepted]{icml2017}
\begin{document}

\twocolumn[
\icmltitle{Learning in Implicit Generative Models}

\icmlsetsymbol{equal}{*}

\begin{icmlauthorlist}
\icmlauthor{Shakir Mohamed}{dm}
\icmlauthor{Balaji Lakshminarayanan}{dm}
\end{icmlauthorlist}

\icmlaffiliation{dm}{DeepMind, London, UK}

\icmlcorrespondingauthor{Shakir Mohamed}{shakir@google.com}
\icmlcorrespondingauthor{Balaji Lakshminarayanan}{balajiln@google.com}

\vskip 0.3in
]

\printAffiliationsAndNotice{\icmlEqualContribution} %

\begin{abstract}
Generative adversarial networks (GANs) provide an algorithmic framework for constructing generative models with several appealing properties: they do not require a likelihood function to be specified, only a generating procedure; they provide samples that are sharp and compelling; and they allow us to harness our knowledge of building highly accurate neural network classifiers. Here, we develop  our understanding of GANs with the aim of forming a rich view of this growing area of machine learning---to build connections to the diverse set of statistical thinking on this topic, of which much can be gained by a mutual exchange of ideas. We frame GANs within the wider landscape of algorithms for learning in implicit generative models---models that only specify a stochastic procedure with which to generate data---and relate these ideas to modelling problems in related fields, such as econometrics and approximate Bayesian computation. We develop likelihood-free inference methods and highlight hypothesis testing as a principle for learning in implicit generative models, using which we are able to derive the objective function used by GANs, and many other related objectives. The testing viewpoint directs our focus to the general problem of density-ratio  and density-difference estimation. There are four approaches for density comparison, one of which is a solution using classifiers to distinguish real from generated data. Other approaches such as divergence minimisation and moment matching have also been explored, and we synthesise these views to form an understanding  in terms of the relationships between them and the wider literature, highlighting avenues for future exploration and cross-pollination.
\end{abstract}

\section{Implicit Generative Models}
It is useful to make a distinction between two types of probabilistic models: prescribed and implicit models \citep{diggle1984monte}. \textit{Prescribed probabilistic models} are those that provide an explicit parametric specification of the distribution of an observed random variable $\vx$, specifying a log-likelihood function $\log q_\theta(\vx)$ with parameters $\vtheta$. Most models in machine learning and statistics are of this form, whether they be state-of-the-art classifiers for object recognition, complex sequence models for machine translation, or fine-grained spatio-temporal models tracking the spread of disease. Alternatively, we can specify \textit{implicit probabilistic models} that  define a stochastic procedure that directly generates data. Such models are the natural approach for problems in climate and weather, population genetics, and ecology, since the mechanistic understanding of such systems can be used to directly create a data simulator, and hence the model. It is exactly because implicit models are more natural for many problems that they are of interest and importance.

Implicit generative models use a latent variable $\vz$ and transform it using a deterministic function $\mathcal{G}_\theta$ that maps from $\mathbb{R}^m \rightarrow \mathbb{R}^d$ using parameters $\vtheta$. Such models are amongst the most fundamental of models, e.g., many of the basic methods for generating non-uniform random variates are based on simple implicit models and one-line transformations \citep{devroye2006nonuniform}. In general, implicit generative models specify a valid density on the output space that forms an effective likelihood function:
\begin{equation}
\vx = \mathcal{G}_{\vtheta}(\vz'); \qquad \vz' \sim \zdist
\end{equation}
\begin{equation}
\qtheta(\vx) = \frac{\partial}{\partial x_1} \ldots \frac{\partial}{\partial x_d} \int_{\{\mathcal{G}_{\vtheta}(\vz) \leq \vx\}} \zdist d \vz, \label{eq:effective_density}
\end{equation}
where $\zdist$ is a latent variable that provides the external source of randomness and equation \eqref{eq:effective_density} is the definition of the transformed density as the derivative of the cumulative distribution function. When the function $\mathcal{G}$ is well-defined, such as when the function is invertible, or has dimensions $m=d$ with easily characterised roots, we recover the familiar rule for transformations of probability distributions. 
We are interested in developing more general and flexible implicit generative models where the function $\mathcal{G}$ is a non-linear function with $d > m$, specified by deep networks. 
The integral \eqref{eq:effective_density} is intractable in this case: we will be unable to determine the set $\{\mathcal{G}_{\vtheta}(\vz) \leq \vx\}$,  the integral will often be unknown even when the integration regions are known and, the derivative is high-dimensional and difficult to compute. Intractability is also a challenge for prescribed models, but the lack of a likelihood term significantly reduces the tools available for learning. In implicit models, this difficulty motivates the need for methods that side-step the intractability of the likelihood \eqref{eq:effective_density}, or are likelihood-free.

Both generative adversarial networks (GANs) \citep{goodfellow2014generative} and classifier ABC \citep{gutmann2014statistical} provide a solution for exactly this type of problem. These approaches specify algorithmic frameworks for learning in implicit generative models, also referred to as generator networks, generative neural samplers or (differentiable) simulator-models. Both approaches rely on a learning principle based on discriminating real from generated data, which we shall show instantiates a core principle of likelihood-free inference, that of hypothesis and two-sample testing. Many of the methods we discuss are known in isolation, spread disparately throughout the literature. \textit{Our core contribution is to make explicit their probabilistic basis and to clearly discuss the connections between them.}

\textbf{Note on notation.} We denote data by the random variable $\vx$, the (unknown) true data density by $\tdist(\vx)$, our (intractable) model density by $\qtheta(\vx)$. $\zdist$ is a density over latent variables $\vz$. Parameters of the model are $\vtheta$, and parameters of the ratio and discriminator functions are $\phi$.
\section{Hypothesis Testing and Density Ratios}
\subsection{Likelihood-free Inference}

Without a likelihood function, many of the widely-used tools for inference and parameter learning become unavailable. But there are tools that remain, including the method-of-moments \citep{hall2005generalized}, the empirical likelihood \citep{owen1988empirical}, Monte Carlo sampling \citep{marin2012approximate}, and mean-shift estimation \citep{fukunaga1975estimation}. Since we can easily draw samples from the model, we can use any method that compares two sets of samples---one from the true data distribution and one from the model distribution---to drive learning. This is a process of \textit{density estimation-by-comparison}, comprising two steps: comparison and estimation. For comparison, we test the hypothesis that the true data distribution $\tdist(\vx)$ and our model distribution $q(\vx)$ are equal, using the \textit{density difference} $r(\vx) = \tdist(\vx) - q(\vx)$, or the \textit{density ratio} $r(\vx) = \sfrac{\tdist(\vx)}{q(\vx)}$. The comparator $r(\vx)$ provides information about the departure of our model distribution from the true distribution and can be motivated by either the Neyman-Pearson lemma or the Bayesian posterior evidence, appearing in the likelihood ratio and the Bayes factor \citep{kass1995bayes}.  For estimation, we use information from the comparison to drive learning of the parameters of the generative model. This is the focus of our approach for likelihood-free inference: estimating density-comparisons and using them as the driving principle for learning in implicit generative models.

\begin{figure*}%
\centering
\includegraphics[width=0.65\textwidth]{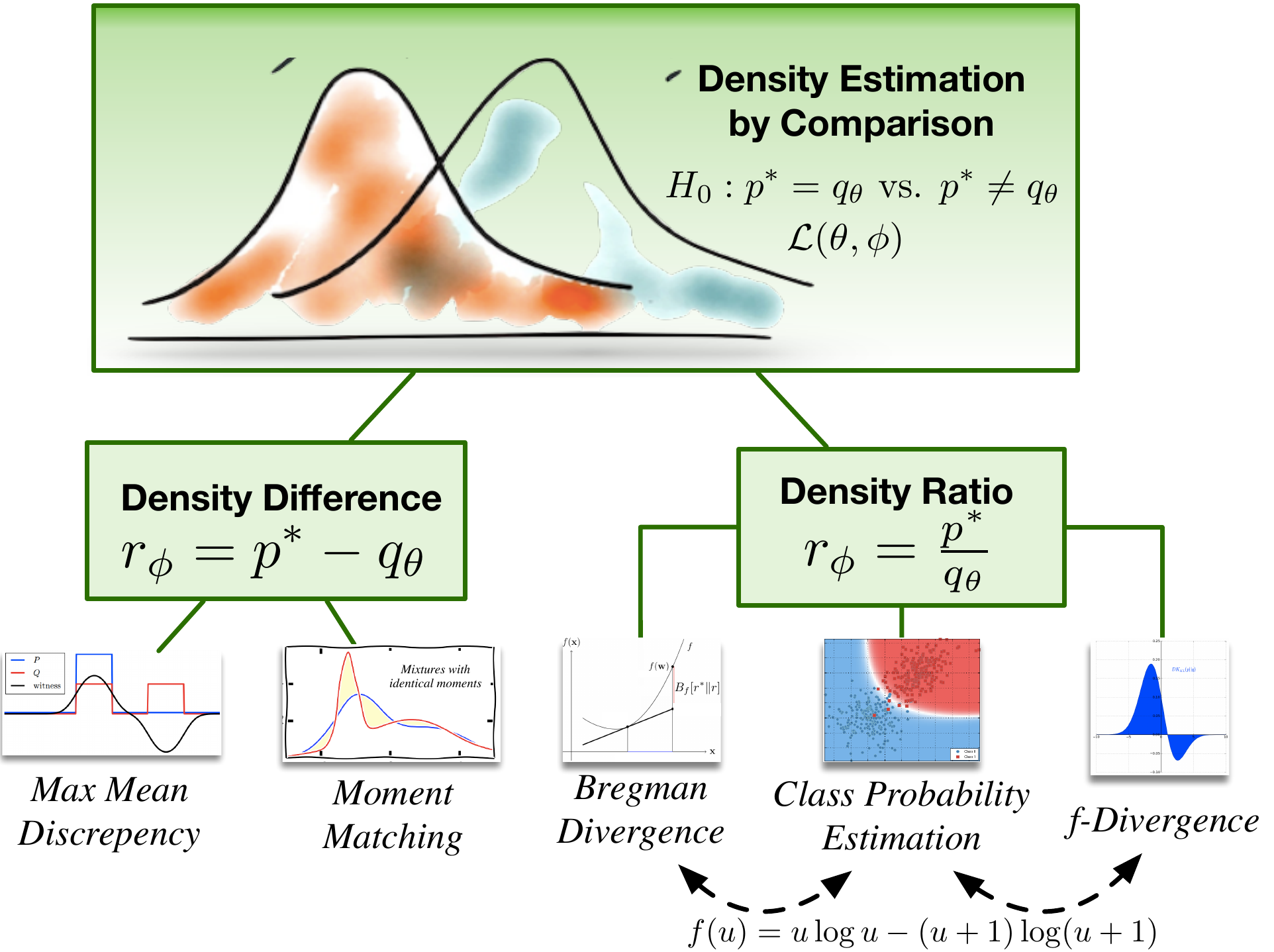}
\caption{Summary of approaches for learning in implicit models. We define a joint function $\mathcal{L}(\phi, \theta)$ and alternate between optimising the loss w.r.t. comparison parameters $\phi$ and model parameters $\theta$.  }
\label{fig:overviewdiag}
\vspace{-4mm}
\end{figure*}

The direct approach of comparing distributions by first computing the individual marginals is not possible with implicit models. By directly estimating the density ratio or difference, and exploiting knowledge of the probabilities involved, it will turn out that comparison can be a much easier problem than computing the marginal likelihoods, and is what will make this approach appealing. There are four general approaches to consider \citep{sugiyama2012density}: 1) class-probability estimation, 2) divergence minimisation, 3) ratio matching, and 4) moment matching. These are highly developed research areas in their own right, but their role in providing a learning principle for density estimation is under-appreciated and opens up an exciting range of approaches for learning in implicit generative models. Figure \ref{fig:overviewdiag} summarises these approaches by showing pathways available for learning, which follow from the choice of inference driven by hypothesis testing and comparison.

\subsection{Class Probability Estimation}
\label{sect:CPE}
The density ratio can be computed by building a classifier to distinguish observed data from that generated by the model. This is the most popular approach for density ratio estimation and the first port of call for learning in implicit models. \citet{hastieElements} call this approach unsupervised-as-supervised learning, \citet{qin1998inferences} explore this for analysis of case-control in observational studies, both \citet{neal2008computing} and \citet{cranmer2015approximating} explore this approach for high-energy physics applications, \citet{gutmann2012noise} exploit it for learning un-normalised models, \citet{lopez2016revisiting} for causal discovery, and \citet{goodfellow2014generative} for learning in implicit generative models specified by neural networks.

We denote the domain of our data by $\mathcal{X} \subset \mathbb{R}^d$. The true data distribution has a density $\tdist(\vx)$ and our model has density $\qtheta(\vx)$, both defined on $\mathcal{X}$. We also have access to a set of $n$ samples $\mathcal{X}_p = \{\vx^{(p)}_1, \ldots, \vx^{(p)}_n\}$ from the true data distribution, and a set of $n'$ samples $\mathcal{X}_q = \{\vx^{(q)}_1, \ldots, \vx^{(q)}_{n'}\}$ from our model.
We introduce a random variable $y$, and assign a label $y= 1$ to all samples in $\mathcal{X}_p$ and $y= 0$ to all samples in $\mathcal{X}_q$. We can now represent $\tdist(\vx) = p(\vx | y = 1)$ and $\qtheta(\vx) = p(\vx | y = 0)$. By application of Bayes' rule, we can compute the ratio $r(\vx)$ as:
\begin{align}
\frac{\tdist(\vx)}{\qtheta(\vx)} & = \frac{p(\vx | y = 1)}{p(\vx | y = 0)} = \left. \frac{p(y=1| \vx)p(\vx)}{p(y = 1)} \middle/ \frac{p(y=0| \vx)p(\vx)}{p(y = 0)} \right. 
\nonumber\\
&= \frac{p(y = 1 |\vx)}{p(y = 0 | \vx)}\cdot\frac{1-\pi}{\pi},
\end{align}
which indicates that the problem of density ratio estimation is equivalent to that of class probability estimation, since the problem is reduced to computing the probability $p(y = 1 | \vx)$. We assume that the marginal probability over classes is $p(y = 1) = \pi$, which allows the relative proportion of data from the two classes to be adjusted if they are imbalanced; in most formulations $\pi = \textrm{\textonehalf}$   for the balanced case, and in imbalanced cases $\tfrac{1-\pi}{\pi} \approx \sfrac{n'}{n}$.

Our task is now to specify a scoring function, or discriminator, \mbox{$\mathcal{D}(\vx; \vphi) = p(\vy = 1| \vx)$}: a function bounded in [0,1] with parameters $\vphi$ that computes the probability of data belonging to the positive (real data) class. This discriminator is related to the density ratio through the mapping $\mathcal{D} = \sfrac{r}{(r+1)}$; $r = \sfrac{\mathcal{D}}{(1-\mathcal{D})}$. Conveniently, we can use our knowledge of building classifiers and specify these functions using deep neural networks. Given this scoring function, we must specify a proper scoring rule \citep{gneiting2007strictly, buja2005loss} for binary discrimination to allow for parameter learning, such as those in Table \ref{tab:scoring_rules}. A natural choice is to use the Bernoulli (logarithmic) loss:
\begin{align}
&\mathcal{L}(\vphi, \vtheta) \nonumber\\
 & = \mathbb{E}_{p(\vx|y)p(y)}[- y\log \mathcal{D}(\vx; \vphi) - (1-y)\log (1 - \mathcal{D}(\vx; \vphi))] \nonumber\\
 & = \pi \mathbb{E}_{\tdist(\vx)}[- \log \mathcal{D}(\vx; \vphi)] \nonumber \\
 & \qquad+ (1-\pi)\mathbb{E}_{\qtheta(\vx)}[- \log (1 - \mathcal{D}(\vx; \vphi))].
\end{align}
Since we know the underlying generative process for $\qtheta(\vx)$, using a change of variables, we can express the loss in term of an expectation over the latent variable $\vz$ and the generative model $\mathcal{G}(\vz; \vtheta)$: 
\begin{align}
\mathcal{L}(\vphi, \vtheta)  &= \pi \mathbb{E}_{\tdist(\vx)}[-\log \mathcal{D}(\vx; \vphi)] \nonumber\\
 &   + (1-\pi)\mathbb{E}_{\zdist}[-\log (1 - \mathcal{D}(\mathcal{G}(\vz; \vtheta); \vphi))]. \label{eq:GAN}
\end{align}

The final form of this objective \eqref{eq:GAN} is exactly that used in generative adversarial networks (GANs) \citep{goodfellow2014generative}. In practice, the expectations are computed by Monte Carlo integration using samples from $\tdist$ and $\qtheta$.
Equation \eqref{eq:GAN} allows us to specify a bi-level optimisation \citep{colson2007overview} by forming a \textit{ratio loss} and a \textit{generative loss}, using which we perform an alternating optimisation. Our convention throughout the paper will be to always form the ratio loss by extracting all terms in $\mathcal{L}$ related to the ratio function parameters $\vphi$, and minimise the resulting objective. For the generative loss, we will similarly extract all terms related to the model parameters $\vtheta$, flip the sign, and minimise the resulting objective. For equation \eqref{eq:GAN}, the bi-level optimisation is:
\begin{align}
&\textrm{\textbf{Ratio loss:}}\,\,
\min_\phi \pi \mathbb{E}_{\tdist(\vx)}[-\log \mathcal{D}(\vx; \vphi)]  \nonumber \\
& \qquad\qquad + (1-\pi)\mathbb{E}_{\qtheta(\vx)}[-\log (1 - \mathcal{D}(\vx; \vphi))] \nonumber \\
&\textrm{\textbf{Generative loss:}} \,\, \min_\theta \mathbb{E}_{\zdist}[\log (1 - \mathcal{D}(\mathcal{G}(\vz; \vtheta)))]. \label{eq:CPE_ratioloss}
\end{align}
The ratio loss is minimised since it acts as a surrogate negative log-likelihood; the generative loss is minimised since we wish to minimise the probability of the negative (generated-data) class. We explicitly write out these two stages to emphasise that the objectives used are separable. While we can derive the generative loss from the ratio loss as we have done, any generative loss that drives $\qtheta$ to $\tdist$, such as minimising the widely-used $\mathbb{E}_{q(\vz)}[-\log \mathcal{D}(\mathcal{G}(\vz; \vtheta))]$ \citep{goodfellow2014generative, nowozin2016f} or $\mathbb{E}_{q(\vz)}[-\log \tfrac{\mathcal{D(\mathcal{G}(\vz; \vtheta))}}{1-\mathcal{D(\mathcal{G}(\vz; \vtheta))}}] = \mathbb{E}_{q(\vz)}[-\log r(\mathcal{G}(\vz; \vtheta))$ \citep{sonderby2016amortised}, is possible. 

Any scoring rule from Table \ref{tab:scoring_rules} can be used to give a loss function for optimisation. These rules are amenable to stochastic approximation and alternating optimisation, as described by \citet{goodfellow2014generative}, along with many of the insights for optimisation that have been developed since \citep{salimans2016improved, radford2015unsupervised, zhao2016energy, sonderby2016amortised}. The Bernoulli loss can be criticised in a number of ways and makes other scoring rules interesting to explore. The Brier loss provides a similar decision rule, and its use in calibrated regression makes it appealing; the motivations behind many of these scoring rules are discussed in \citep{gneiting2007strictly}. 
Finally, while we have focussed on the two sample hypothesis test, we believe it could be advantageous to extend this reasoning to the case of multiple testing, where we simultaneously test several sets of data \citep{bickel2008multi}.
\begin{table*}[t]
\centering
\begin{tabular}{l|l}
\hline
\textbf{Loss} & \textbf{Objective Function} ($\mathcal{D} := \mathcal{D}(\vx; \vphi)$)\\
\hline \hline
Bernoulli loss & $\pi \mathbb{E}_{\tdist(\vx)}[-\log \mathcal{D}] + (1-\pi)\mathbb{E}_{\qtheta(\vx)}[-\log (1 - \mathcal{D})]$ \\
Brier score & $\pi \mathbb{E}_{\tdist(\vx)}[(1- \mathcal{D})^2] + (1-\pi)\mathbb{E}_{\qtheta(\vx)}[\mathcal{D}^2]$ \\
Exponential loss & $\pi \mathbb{E}_{\tdist(\vx)}\left[\left( \frac{1 - \mathcal{D}}{\mathcal{D}} \right)^{\frac{1}{2}}\right] + (1-\pi)\mathbb{E}_{\qtheta(\vx)}\left[\left( \frac{\mathcal{D}}{1-\mathcal{D}} \right)^{\frac{1}{2}}\right]$ \\
Misclassification  & $\pi \mathbb{E}_{\tdist(\vx)}[\mathbb{I}[\mathcal{D} \leq 0.5]] + (1-\pi)\mathbb{E}_{\qtheta(\vx)}[\mathbb{I}[\mathcal{D} > 0.5]]$ \\
Hinge loss & $ \pi \mathbb{E}_{\tdist(\vx)}\left[\max\left(0, 1 - \log \tfrac{\mathcal{D}}{1-\mathcal{D}}\right)\right] + (1-\pi)\mathbb{E}_{\qtheta(\vx)}\left[\max\left(0, 1+  \log \tfrac{\mathcal{D}}{1-\mathcal{D}}\right)\right]$\\
Spherical & $ \pi  \mathbb{E}_{\tdist(\vx)}\left[-\alpha\mathcal{D}\right] + (1-\pi)\mathbb{E}_{\qtheta(\vx)}\left[-\alpha(1- \mathcal{D})\right]; \quad \alpha = (1 - 2\mathcal{D} + 2\mathcal{D}^2)^{-\sfrac{1}{2}}$  \\
\hline
\end{tabular}
\caption{Proper scoring rules that can be minimised in class probability-based learning of implicit generative models.}
\label{tab:scoring_rules}
\vspace{-4mm}
\end{table*}

The advantage of using a proper scoring rule is that the global optimum is achieved iff $\qtheta=\tdist$ (cf.~the proof in \citep{goodfellow2014generative} for the Bernoulli loss); however there are no convergence guarantees since the optimisation is non-convex.  
\citet{goodfellow2014generative} discuss the relationship to maximum likelihood estimation, which minimises the divergence $\KL[p^* \| q]$, and show that the GAN objective with Bernoulli loss is instead related to the Jensen Shannon divergence $\JSpq{\tdist}{q}$.
In the objective \eqref{eq:GAN}, $\pi$ denotes the marginal probability of the positive class; however several authors have proposed choosing $\pi$  depending on the problem. In particular, \citet{huszar2015not} showed that varying $\pi$ is related to optimizing a generalised Jensen-Shannon divergence $\JSpqpi{\tdist}{q}$. \citet{creswell2016task} presented results showing that different values of $\pi$ are desirable, depending on whether we wish to fit one of the modes (a `high precision, low recall' task such as generation) or explain all of the modes (a `high recall, low precision' task such as retrieval).

\subsection{Divergence Minimisation }
A second approach to testing is to use the divergence between the true density $\tdist$ and our model $q$, and use this as an objective to drive learning of the generative model. A natural class of divergences to use are the $f$-divergences (or Ali-Silvey \citep{ali1966general} or Csiszar's $\phi$-divergence \citep{csisz1967information}) since they are fundamentally linked to the problem of two-sample hypothesis testing \citep{liese2008f}: $f$-divergences represent an integrated Bayes risk since they are an expectation of the density ratio. \citet{nowozin2016f} develop $f$-GANs using this view. The $f$-divergences contain the KL divergence as a special case and are equipped with an exploitable variational formulation:
\begin{align}
&D_f\left[\tdist(\vx)\| \qtheta(\vx)\right] = \int \qtheta(\vx) f\left(\frac{\tdist(\vx)}{\qtheta(\vx)}\right) d\vx  \nonumber\\
&\qquad= \mathbb{E}_{\qtheta(\vx)}[f(r(\vx))]
\nonumber\\
&\qquad\geq \sup_t \mathbb{E}_{\tdist(\vx)}[t(\vx)] - \mathbb{E}_{\qtheta(\vx)}[f^\dag(t(\vx))] \label{eq:f-divergence}
\end{align}
where $f$ is a convex function with derivative $f'$ and Fenchel conjugate $f^\dag$; this divergence class instantiates many familiar divergences, such as the KL and Jensen-Shannon divergence. 
The variational formulation introduces the functions $t(\vx)$ whose optimum is related to the density ratio since $t^*(\vx) = f'(r(\vx))$. Substituting $t^*$ in \eqref{eq:f-divergence}, we transform the objective \eqref{eq:fdiv_obj} into supremum over $r_\phi$ (which is attained when $r_\phi=r^*=\sfrac{\tdist}{\qtheta}$). For self-consistency, we flip the sign to make it a minimisation problem in $r_\phi$, leading to the bi-level optimisation:
\begin{eqnarray}
 \mathcal{L} = \mathbb{E}_{\tdist(\vx)}[-f'(r_\phi(\vx))] + \mathbb{E}_{\qtheta(\vx)}[f^\dag(f'(r_\phi(\vx))] \label{eq:fdiv_obj}
\end{eqnarray}
\begin{align}
& \textrm{\textbf{Ratio loss:}} \nonumber \\ 
& \min_\phi \mathbb{E}_{\tdist(\vx)}[-f'(r_\phi(\vx))] + \mathbb{E}_{\qtheta(\vx)}[f^\dag(f'(r_\phi(\vx))] \label{eq:fdiv_ratioloss}\\
& \textrm{\textbf{Generative loss:}} \,\, \min_\theta \mathbb{E}_{\zdist}[-f^\dag(f'(r(\mathcal{G}(\vz; \vtheta)))], \label{eq:fdiv_genloss}
\end{align}
where we derived equation \eqref{eq:fdiv_genloss} by extracting all the terms involving $\qtheta(\vx)$ in equation \eqref{eq:fdiv_obj}, used the change of variables to express it in terms of the underlying generative model and flipping the sign to obtain a minimisation. There is no discriminator in this formulation, and this role is taken by the ratio function. We minimise the ratio loss, since we wish to minimise the negative of the variational lower bound; we minimise the generative loss since we wish to drive the ratio to one. By using the function $f(u) = u \log u$, we recover an objective using the KL divergence; when $f(u) = u \log u - (u+1)\log(u+1)$, we recover the objective function in equation \eqref{eq:GAN} from the previous section using the Bernoulli loss, and hence the objective for GANs.

The density ratio implies that $\tdist(\vx) \approx \tilde{p} = r_\phi(\vx)\qtheta(\vx)$, since it is the amount by which we must correct our model $\qtheta(\vx)$ to match the true distribution.  This led us to a divergence formulation that evaluates the divergence between the distributions $\tdist$ and $\tilde{p}$, using the KL divergence:
\begin{align}
 D_{KL}[\tdist(\vx) \| \tilde{p}(\vx)]  = & \int \tdist(\vx) \log \frac{\tdist(\vx)}{r_\phi(\vx)\qtheta(\vx)}d\vx \nonumber\\
&+ \int (r_\phi(\vx)\qtheta(\vx) - \tdist(\vx)) d\vx
\end{align}
\begin{align}
 \mathcal{L}  & =  \mathbb{E}_{\tdist(\vx)}[- \log r_\phi(\vx) ] + \mathbb{E}_{\qtheta(\vx)} [r_\phi(\vx)-1] \nonumber\\
& \quad - \mathbb{E}_{\tdist
(\vx)}[\log \qtheta(\vx)] + \mathbb{E}_{\tdist(\vx)}[\log \tdist(\vx)],  \label{eq:KLIEP}
\end{align}
where the first equation is the KL for un-normalised densities \citep{minka2005divergence}.
This leads to a convenient and valid ratio loss since all terms independent of $r$ can be ignored. 
 \citet{sugiyama2012densitybook} refer to this objective as KL importance estimation procedure (KLIEP).
But we are unable to derive a useful generative loss from this expression since the third term with $\log q$ in \eqref{eq:KLIEP}  cannot be ignored, and is unavailable for implicit models. Since the generative loss and ratio losses need not be coupled, any other generative loss can be used, e.g., equation \eqref{eq:CPE_ratioloss}. But this is not ideal, since we would prefer to derive valid ratio losses from the same principle to make reasoning about correctness and optimality easier. We include this discussion to point out that while the formulation of equation \eqref{eq:f-divergence} is generally applicable, the formulation \eqref{eq:KLIEP}, while useful for ratio estimation, is not useful for generative learning. The difficulty of specifying stable and correct generative losses is a common theme of work in GANs; we will see a similar difficulty in the next section.

The equivalence between divergence minimisation and class probability estimation is also widely discussed, and most notably developed by \citet{reid2011information} using the tools of weighted integral measures, and more recently by \citet{menon2016linking}. This divergence minimisation viewpoint \eqref{eq:f-divergence} was used in $f$-GANs and explored in depth by \citet{nowozin2016f} who provide a detailed description, and explore many of the objectives that become available and practical guidance. 

\subsection{Ratio matching}
A third approach is to directly minimise the error between the true density ratio and an estimate of it. 
Denoting the true density ratio as $r^*(\vx) = \sfrac{\tdist(\vx)}{\qtheta(\vx)}$ and its approximation as $r_\phi(\vx)$, we can define a loss using the squared error:
\begin{align}
\mathcal{L} & = \frac{1}{2}\int \qtheta(\vx) (r(\vx)- r^*(\vx))^2 d\vx \label{eq:ratio_min_sq}
 \\
&= \tfrac{1}{2} \E_{\qtheta(\vx)}[r_\phi(\vx)^2] - \E_{\tdist(\vx)}[r_\phi(\vx)] + \tfrac{1}{2}\E_{\tdist(\vx)}[r^*(\vx)]  \nonumber\\
& = \tfrac{1}{2} \E_{\qtheta(\vx)}[r_\phi(\vx)^2] - \E_{\tdist(\vx)}[r_\phi(\vx)] \quad s.t.\,\,  r_\phi(\vx) \geq 0, \nonumber
\end{align}
where the final objective is obtained by ignoring terms independent of $r_\phi(\vx)$ and is used to derive ratio and generative losses. 
When used to learn the ratio function, \citet{sugiyama2012densitybook} refer to this objective as 
least squares importance fitting (LSIF).
Concurrently with this work, \citet{uehara2016ratio} recognised the centrality of the density ratio, the approach for learning by ratio matching, and its connections to GANs \citep{goodfellow2014generative} and $f$-GANs \citep{nowozin2016f}, and provide useful guidance on practical use of ratio matching.

We can generalise \eqref{eq:ratio_min_sq} to loss functions beyond the squared error using the Bregman divergence for density ratio estimation \citep{sugiyama2012density, uehara2016ratio}, and is the unifying tool exploited in previous work \citep{reid2011information,sriperumbudur2009integral, sugiyama2012density, menon2016linking}. This leads to a minimisation of the Bregman divergence $B_f$ between ratios:
\begin{align}
&B_f(\rstarx\|\rphix) \nonumber\\ 
& = \E_{\qtheta(\vx)} \bigl(f(\rstarx) -f(\rphix)  \nonumber\\
& \qquad - f'(\rphix)\bigl[\rstarx-\rphix\bigr]\bigr) \label{eq:bregdiv}\\
& = \E_{\qtheta(\vx)}\left[\rphix f'(\rphix)-f(\rphix)\right]   \nonumber\\
&\qquad - \E_{\tdist}[f'(\rphix)] + D_f[\tdistx\|\qthetax] \label{eq:bregdiv1} \\
&= \LBregman(\rphix) + D_f[\tdistx\|\qthetax], \label{eq:bregdiv2}
\end{align}
where we have used $\tdist = r^*\qtheta$, and $D_f$ is the $f$-divergence defined in equation \eqref{eq:f-divergence}. We can derive a ratio loss from \eqref{eq:bregdiv2} by extracting all the terms in $r_\phi$, leading to the minimisation of $\LBregman(r_\phi)$.  
The role of the discriminator in GANs is again taken by the ratio $r$ that provides information about the relationship between the two distributions.
This ratio loss, like in \citet{uehara2016ratio}, is \textit{equivalent} to the ratio loss we derived using divergence minimisation \eqref{eq:fdiv_obj}, since:
\begin{align}
&\LBregman(\rphi(\vx))  \nonumber\\
&
= E_{\tdist}[-f'(\rphix)]  \nonumber\\
&\quad + \E_{\qtheta(\vx)}[\rphix f'(\rphix)-f(\rphix)]  \label{eq:lbreg_bdiv}\\
&= E_{\tdist}[-f'(\rphix)]  + \E_{\qtheta(\vx)}[f^\dag(f'(\rphix))]. \label{eq:lbreg_fdiv}
 \end{align}
The equivalence of the second terms in \eqref{eq:lbreg_bdiv} and \eqref{eq:lbreg_fdiv} can be derived by using the definition of the dual function: 
\begin{align}
f^\dag(f'(x)) = \max_r \ r f'(x) - f(r).  
\end{align}
The maximum is attained at $x=r$, leading to the identity 
\begin{align}
f^\dag(f'(\rphix))=\rphix f'(\rphix)-f(\rphix). 
\end{align}

If we follow the strategy we used in previous sections to obtain a generative loss, by collecting the terms in equation \eqref{eq:bregdiv2} dependent on $\qtheta$, we obtain:
\begin{align}
\mathcal{L}(\qtheta) =&  \E_{\qtheta(\vx)}[\rphix f'(\rphix)] - \E_{\qtheta(\vx)}[f(\rphix)]  \nonumber\\
& + D_f[\tdistx||\qthetax]. \label{eq:bregmangenloss}
 \end{align}
But this does not lead to a useful generative loss since there are terms involving $q_\theta$ whose density is always unknown, similar to the difficulty we encountered with divergence minimisation in equation \eqref{eq:KLIEP}. Since we can use any generative loss that drives the ratio to one, we can employ other losses, like the alternatives described in section \ref{sect:CPE} . One approximation suggested by \citet{uehara2016ratio} is to assume $\tdist\approx \rphi\qtheta$, i.e. assume a near-optimal ratio, which reduces the $f$-divergence to:
\begin{align}
&\!D_f[\tdistx\|\qthetax] \! =\! \E_{\qtheta(\vx)}\left[f\left(\frac{\tdist}{\qthetax}\right)\right] \nonumber\\
&\approx  \E_{\qtheta(\vx)}\left[f\left(\frac{\qthetax\rphix}{\qthetax}\right)\right] = \E_{\qtheta(\vx)}[f(\rphix)]
 \end{align}
As a result, the last two terms in equation \eqref{eq:bregmangenloss} cancel, leaving a generative loss that can be used for learning. Using ratio matching under the Bregman divergence we obtain the bi-level optimisation:
\begin{align}
&\textrm{\textbf{Ratio loss:}}\,\, \nonumber\\
& \min_\phi \E_{\qtheta(\vx)}[\rphix f'(\rphix)-f(\rphix)]  - \E_{\tdist}[f'(\rphix)] \nonumber %
\\
& \textrm{\textbf{Generative loss:}} \,\, \min_\theta \E_{\qtheta(\vx)}[\rphix f'(\rphix)]  \label{eq:bdiv_genloss}
\end{align}

\subsection{Moment Matching}
A final approach for testing is to evaluate whether the moments of the distributions $\tdist$ and $q$ are the same, i.e. by moment matching. We compare the moments of the two distributions by minimising their distance, using test statistics $s(\vx)$ that provides the moments of interest:
\begin{align}
\mathcal{L}(\vphi, \vtheta) & = (\mathbb{E}_{\tdist(\vx)}[s(\vx)] - \mathbb{E}_{\qtheta(\vx)}[s(\vx)])^2 \nonumber\\
& = (\mathbb{E}_{\tdist(\vx)}[s(\vx)] - \mathbb{E}_{\zdist}[s(\gen)])^2 \label{eq:moment_match_direct}
\end{align}
The choice of test statistics $s(\vx)$ is critical, since ideally, we wish to match all moments of the two distributions.  When the functions $s(\vx)$ are defined within a reproducing kernel Hilbert space, we obtain kernel-based forms of these objectives, which are highly flexible and allow easy handling of data such as images, strings and graphs. The objective \eqref{eq:moment_match_direct} can then be re-expressed in closed-form in terms of kernel functions, leading to the maximum mean discrepancy criterion (MMD). The role of the MMD for learning in implicit generative models defined by neural networks has been explored by both \citet{li2015generative} and \citet{dziugaite2015training}. While typically expensive to compute, there are now highly efficient approaches for computing the MMD \citep{chwialkowski2015fast}. The objective using feature matching by \citet{salimans2016improved} is a form of moment matching that defines the test statistics using intermediate layers of the discriminator function. And a further set of objective functions is possible using other parameterisations of the density difference \citep{sugiyama2013density}. 

The other significant body of research that focusses on learning in implicit generative models using the lens of moment matching is approximate Bayesian computation (ABC) \citep{marin2012approximate}. The models in ABC are often related to problems in population genetics and ecology, where knowledge of these systems leads to ABC being easily exploited to learn simulator parameters. The ABC literature has widely-explored the use of  test statistics, including fixed functions and kernels, and Markov chain Monte Carlo methods to learn the posterior distribution of the parameters. There is a great deal of opportunity for exchange between GANs, ABC and ratio estimation in aspects of scalability, applications, and theoretical understanding. A further insight obtained from a moment-matching approach is that other empirical measures such as the Dudley and Wasserstein distances can easily be considered, establishing even further connections to other growing areas, such as optimal transport \citep{frogner2015learning, sriperumbudur2009integral}; \citet{wgan} has since demonstrated the effectiveness of using the Wasserstein  distance. 

The moment matching approach can be generalised by representing them as an integral probability metric \citep{sriperumbudur2009integral}, which is what makes it possible to describe the relationships and ways of transforming from density ratio estimation using using moment-matching, $f$-divergences, and class-probability estimation. \citet{sriperumbudur2009integral} showed that $f$-divergences and integral probability metrics (MMD, Wasserstein) intersect only at the total variation distance.
As we described previously, by cleverly formulating our problem as a Bregman divergence minimisation, we show that all the methods we described for density ratio estimation are very closely related \citep{sugiyama2012density, reid2011information, sriperumbudur2009integral, sugiyama2012densitybook, uehara2016ratio}.
\section{Choice of Loss Functions}
\label{sect:loss_choice}
Learning in implicit generative models always involves two steps, a comparison step for which we used the density ratio or difference estimators, and an estimation step to learn the parameters of the generative models. The literature on density ratio estimation addresses only the first of these tasks: where \textit{only} the density ratio is to be estimated, \citet{sugiyama2012density} provide guidance on this choice. But when we learn implicit generative models,  we have two loss functions, and these loss functions need not be coupled to each other. \citet{pooleimproved} independently proposed using different $f$ divergences for ratio loss and the generator loss.

\textbf{Evaluation.}
While natural to ask which loss function should be used, this choice is not clear due to the challenges in evaluating implicit models. In prescribed models, the standard approach for evaluation involves computation of the marginalised likelihood, but is not generally possible in implicit models. Estimates of the likelihood can be obtained using kernel density estimators, but this is highly-unreliable, especially in high-dimensions \citep{theis2015note}. Most papers rely on visual inspection that can be misleading, since mode-collapse (where the generated samples sample from only a few modes) and memorisation of the training data cannot be detected easily. Other approaches for evaluation include reporting the value of the density ratio, a quantity we can always compute using the approaches described, the use of annealed importance sampling \citet{wu2016quantitative}, use of empirical distance metric such as the MMD \citep{sutherland2016generative} or the Wasserstein distance \citep{wgan}. But we still lack the tools to make theoretical statements that allow us to assess the correctness of the model learning framework, although theoretical developments in the literature on approximate Bayesian computation may help in this regard, e.g., \citet{frazier2016asymptotic}.

\begin{figure}%
\centering
\includegraphics[width=0.85\columnwidth]{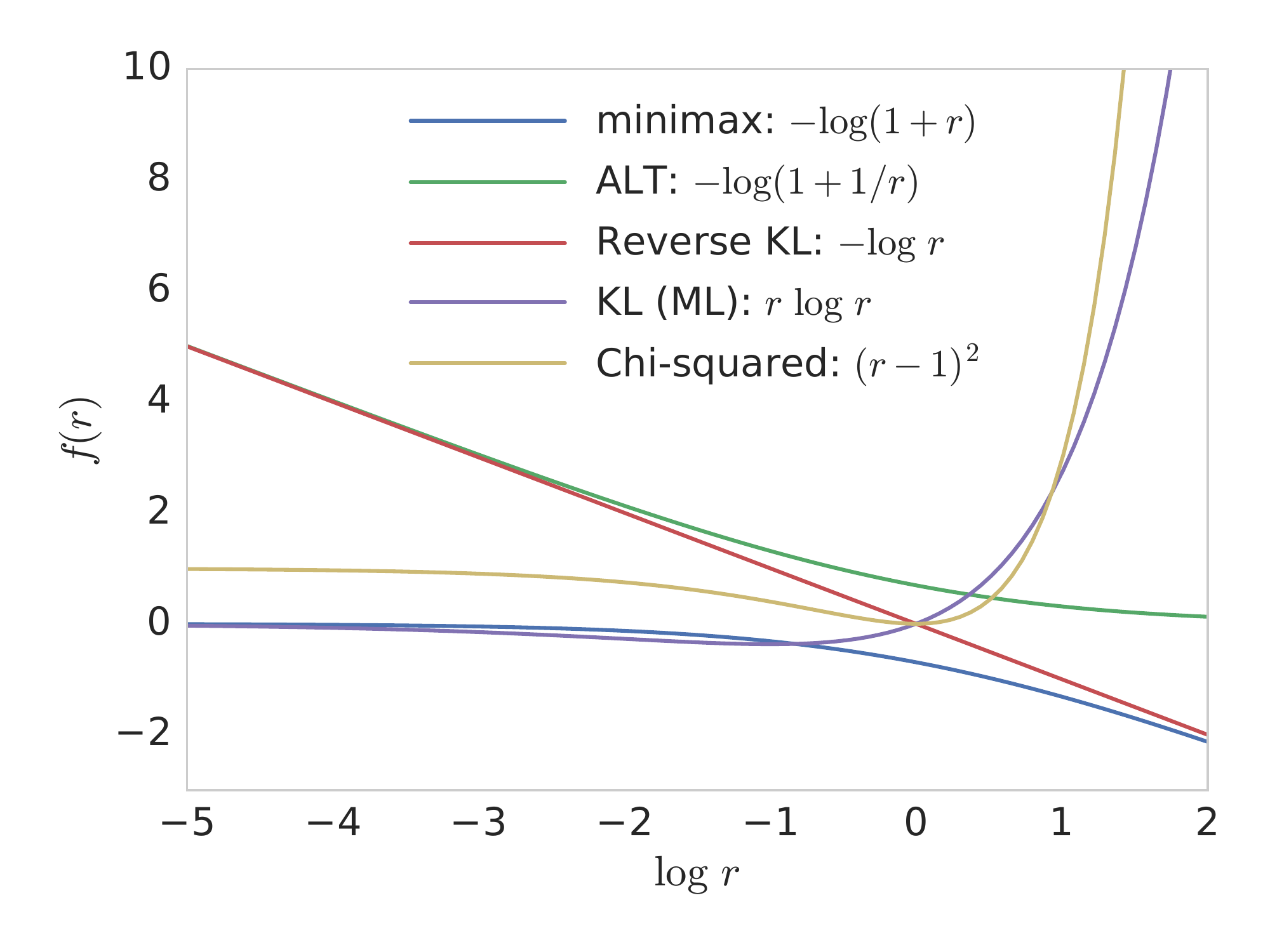}
\vspace{-6mm}
\caption{Objective functions for different choices of $f$.}
\label{fig:genobj}
\vspace{-6.5mm}
\end{figure}

\textbf{Training considerations.}
As discussed, density ratio matching gives us guidance only on the choice of ratio loss. In implicit models, we not only require a good approximation for the density ratio, but also need to ensure that we are able to train the generator efficiently. A meaningful $f$ divergence requires the support of the distributions to overlap and  it is common to add \emph{instance noise} \citep{sonderby2016amortised, arjovsky2017towards} to ensure this.  Additionally, when using gradient based methods, we need to ensure that the generator gradients do not vanish. 
For simplicity, consider the scenario where the implicit model is learnt using the (approximate) $f$-divergence $\E_{\vx\sim\qthetax} f(\rstarx) \approx \E_{\vx\sim\qthetax} f(\rphix)$ using gradient descent. We visualise $f(r)$ for several choices of $f$ in Figure~\ref{fig:genobj}.  
During the initial phase of training, $r$ is very close to $0$ for samples generated from $q$. When optimising the model parameters $\vtheta$, we require that the gradient $f'(r)$ be non-zero for $r\ll 0$. We observe that the $f$ corresponding to KL, chi-squared and minimax is fairly flat when $\log r <0$ and hence difficult to train. %
 For this reason, \citet{goodfellow2014generative} use an alternative loss when training the generator, but which has the same fixed points. We observe that $f$ corresponding to the alternative loss as well as the reverse KL provide stronger gradients when $\log r<0$. 
Similar phenomena are observed when optimising the model according to \eqref{eq:fdiv_genloss} and \eqref{eq:bdiv_genloss}. 
Hence, $f$-GANs \citep{nowozin2016f} and $b$-GANs \citep{uehara2016ratio} optimise alternative losses (with same fixed points) which provide stronger gradients.  
Moment matching methods do not suffer from the aforementioned vanishing-gradient issue. However, they are computationally more expensive than training a classifier,
 and may require larger batch sizes to ensure that the moments are approximated accurately.  
Wasserstein GANs \citep{wgan} are promising as they solve the vanishing-gradient issue in a computationally efficient manner. 

\section{Discussion}
By using an inferential principle driven by hypothesis testing, we have been able to develop a number of indirect methods for learning the parameters of generative models. These methods do not compute the probability of the data or posterior distributions over latent variables, but instead only involve relative statements of probability by comparing populations of data from the generative model to observed data. This view allows us to better understand how algorithms such as generative adversarial networks, approximate Bayesian computation, noise-contrastive estimation, and density ratio estimation are related. Ultimately, these techniques make it possible for us to make contributions to applications in climate and weather, economics, population genetics, and epidemiology, all areas whose principal tools are implicit generative models. 

\textbf{Distinction between implicit and prescribed models.}
The distinction between implicit and prescribed models is useful to keep in mind for at least two reasons: the choice of model has direct implications on the types of learning and inferential principles that can be called upon; and it makes explicit that there are many different ways in which to specify a model that captures our beliefs about data generating processes. 
Any implicit model can be easily turned into a prescribed model by adding a simple likelihood function (noise model) on the generated outputs, so the distinction is not essential. And models with likelihood functions also regularly face the problem of intractable marginal likelihoods. 
But the specification of a likelihood function provides knowledge of $\tdist$ that leads to different algorithms by exploiting this knowledge, e.g., NCE resulting from class-probability based testing in un-normalised models \citep{gutmann2012noise}, or variational lower bounds for directed graphical models. We strive to maintain a clear distinction between the choice of model, choice of inference, and the resulting algorithm, since it is through such a structured view that we can best recognise the connections between research areas that rely on the same sets of tools.

\textbf{Model misspecification and non-maximum likelihood methods.}
Once we have made the choice of an implicit generative model, we cannot use likelihood-based techniques, which then makes testing and estimation-by-comparison appealing. What is striking, is that this leads us to principles for parameter learning that do not require inference of any underlying latent variables, side-stepping one of the major challenges in statistical practice. This piques our interest in more general approaches for non-maximum likelihood and likelihood-free estimation methods, of which there is much work \citep{lyu2011unifying, gutmann2012noise, hall2005generalized, marin2012approximate, frogner2015learning}. We often deal with misspecified models where $\qtheta$ cannot represent $\tdist$, and non maximum likelihood methods could be a more robust choice depending on the task (see figure~1 in \citep{huszar2015not} for an illustrative example).

\textbf{Bayesian inference and message passing.} We have mainly discussed point estimation methods for parameter learning. It is also desirable to perform Bayesian inference in implicit models, where we learn the posterior distribution over the model parameters $p(\vtheta | \vx)$, allowing knowledge of parameter uncertainty to be used in risk minimisation and other decision-making tasks. This is the aim of approximate Bayesian computation (ABC) \citep{marin2012approximate}. The most common approach for thinking about ABC is through moment-matching, but as we explored, there are other approaches available. An approach through class-probability estimation is appealing and leads to classifier ABC \citep{gutmann2014statistical}. We have highly diverse approaches for Bayesian reasoning in prescribed models, and it is desirable to develop a similar breadth of choice for implicit models. 

Implicit models allow for a natural approach for amortised inference, and can be used whenever we wish to to learn distributions from which we do not wish to evaluate probabilities but only generate samples. Consequently, wherever we see  density-ratios or density-differences in probabilistic modelling, we can make use of implicit models and bi-level optimisation, such as in importance sampling, variational inference, or message passing. For example,  \citet{adversarialmp} use GAN-like techniques for inference in factor graphs, and since the central quantity of variational inference is a density ratio it is possible to propose a modified variational inference using implicit models, as discussed by \citet{adversarialvb} and \citet{huszar2017avae}.

\textbf{Perceptual losses.}  
Several authors have also proposed using pre-trained discriminative networks to define the test functions since the difference in activations (of say a pre-trained VGG classifier) can better capture perceptual similarity than the reconstruction error in pixel space. This provides a strong motivation for
further research into \textit{joint} models of images and labels. However, it is not completely unsupervised as the pre-trained discriminative network contains information about labels and invariances. 
This makes evaluation difficult since we lack good metrics and can only fairly compare to other joint models that use both label and image information.

\textbf{Non-differentiable models.}
We have restricted our development to implicit models that are differentiable. In many practical applications, the implicit model (or simulator) will be non-differentiable, discrete or defined in other ways, such as through a stochastic differential equation . 
The stochastic optimisation problem we are generally faced with (for differentiable and non-differentiable models), is to compute $\Delta = \nabla_\theta \mathbb{E}_{q_\theta(\vx)}[f(\vx)]$, the gradient of the expectation of a function. As our exposition followed, when the implicit model is differentiable, the pathwise derivative estimator can be used, i.e. $\Delta =  \mathbb{E}_{q(\vz)}[\nabla_\theta f(\mathcal{G}_\theta(\vz))]$ by rewriting the expectation in terms of the known and easy to sample distribution $q(\vz)$. It is commonly assumed that when we encounter non-differentiable functions that the score function estimator (or likelihood ratio or reinforce estimator) can be used; the score-function estimator is $\Delta = \mathbb{E}_{q_\theta(\vx)}[ f(\vx) \nabla_\theta \log q_\theta(\vx)]$. For implicit models, we do not have knowledge of the density $q(\vx)$ whose log derivative we require, making this estimator inapplicable. This leads to the first of three tools available for non-differentiable models: \textit{weak derivative and related stochastic finite difference} estimators, which require forward-simulation only and compute gradients by perturbation of the parameters \citep{glasserman2003monte, fu2005stochastic}. 

The two other approaches are: \textit{moment matching and ABC-MCMC} \citep{marjoram2003markov}, which has been successful for many problems with moderate dimension; and the natural choice of  \textit{gradient-free optimisation} methods, which include familiar tools such as Bayesian optimisation \citep{gutmann2016}, evolutionary search like CMA-ES, and the Nelder-Mead method, amongst others \citep{conn2009introduction}.
For all three approaches, new insights will be needed to help scale to high-dimensional data with complex dependency structures.

\textit{Ultimately, these concerns serve to highlight the many opportunities that remain for advancing our understanding of inference and parameter learning in implicit generative models.}

\subsubsection*{Acknowledgements}
We thank 
David Pfau, Lars Buesing, Guillaume Desjardins, Theophane Weber, Danilo Rezende, Charles Blundell, Irina Higgins, Yee Whye Teh, Avraham Ruderman,  Brendan O'Donoghue and Ivo Danihelka 
for helpful feedback and discussions. 
We thank Cheng Soon Ong for an insightful conversation that helped to make an initial connection.

\balance
\setlength{\bibsep}{4pt}
\bibliographystyle{abbrvnat}
\bibliography{gan}

\end{document}